\pdfoutput=1

\documentclass[11pt]{article}

\usepackage[final]{acl}

\usepackage{times}
\usepackage{latexsym}
\usepackage{array}
\usepackage{booktabs}
\usepackage{graphicx}
\usepackage{multirow}
\usepackage{todonotes}

\usepackage[T1]{fontenc}

\usepackage[utf8]{inputenc}

\usepackage[T2A]{fontenc}  
\usepackage[utf8]{inputenc}  
\usepackage[english]{babel}  

\usepackage{microtype}

\usepackage{inconsolata}

\usepackage{graphicx}

\usepackage[T2A]{fontenc}  
\usepackage[utf8]{inputenc}  

\usepackage{tcolorbox}
\usepackage{geometry}
\usepackage{times}
\usepackage{tabu}
\usepackage{latexsym}
\usepackage{stmaryrd}
\usepackage{amsmath}
\usepackage{multirow}
\usepackage{bbm}
\usepackage{graphicx}
\usepackage{amssymb}
\usepackage{booktabs}
\usepackage{algpseudocode}
\usepackage{algorithm}
\usepackage{booktabs}
\usepackage{graphicx}
\usepackage{caption}
\usepackage{subcaption}
\usepackage{tabularx}
\usepackage{colortbl}
\usepackage{xcolor}
\usepackage{tablefootnote}
\usepackage{enumitem}  
\usepackage{lipsum}
\usepackage{xspace}
\usepackage{bbding}
\usepackage{pifont}
\usepackage{arydshln}
\usepackage{array}
\usepackage{cleveref}
\usepackage{xspace}
\usepackage{url}
\newcommand{\PreserveBackslash}[1]{\let\temp=\\#1\let\\=\temp}
\newcolumntype{C}[1]{>{\PreserveBackslash\centering}p{#1}}
\newcolumntype{R}[1]{>{\PreserveBackslash\raggedleft}p{#1}}
\newcolumntype{L}[1]{>{\PreserveBackslash\raggedright}p{#1}}

\usepackage{hyperref}

\def\xHyphenate#1#2\wholeString {\if#1%
    \else\transform{#1}%
    \takeTheRest#2\ofTheString\fi}
\def\takeTheRest#1\ofTheString\fi
{\fi \xHyphenate#1\wholeString}
\def\transform#1{\url{#1}\hskip 0pt plus 1pt}

\usepackage[T1]{fontenc}

\usepackage{microtype}

\newcommand{\datasetname}{\texttt{KazMMLU}\xspace}

\definecolor{mygreen}{RGB}{217, 234, 211}
\definecolor{myred}{RGB}{244, 204, 204}

\title{\datasetname: Evaluating Language Models on Kazakh, Russian, \\and Regional Knowledge of Kazakhstan}

\author{Mukhammed Togmanov$^1$\thanks{These authors contributed equally.}\quad Nurdaulet Mukhituly$^1$$^*$ \quad  Diana Turmakhan$^1$$^*$ \\  \textbf{Jonibek Mansurov}$^1$ \quad   \textbf{Maiya Goloburda}$^1$ \quad    \textbf{Akhmed Sakip}$^1$ \quad   \textbf{Zhuohan Xie}$^1$ \\   \textbf{Yuxia Wang}$^1$  \quad   \textbf{Bekassyl Syzdykov}$^1$  \quad  \textbf{Nurkhan Laiyk}$^1$ \quad  \textbf{Alham Fikri Aji}$^1$  \\ \textbf{Ekaterina Kochmar}$^1$ \quad     \textbf{Preslav Nakov}$^{1,2}$  \quad  \textbf{Fajri Koto}$^1$ \\
$^1$Mohamed bin Zayed University of Artificial Intelligence \\
$^2$Institute of Foundation Models\\
	\texttt{\small \{mukhammed.togmanov,nurdaulet.mukhituly,diana.turmakhan\}@mbzuai.ac.ae 
	} 
}

\begin{document}
\maketitle

\begin{abstract}
Despite having a population of twenty million, Kazakhstan's culture and language remain underrepresented in the field of natural language processing. Although large language models (LLMs) continue to advance worldwide, progress in Kazakh language has been limited, as seen in the scarcity of dedicated models and benchmark evaluations. To address this gap, we introduce \datasetname, the first MMLU-style dataset specifically designed for Kazakh language. \datasetname comprises 23,000 questions that cover various educational levels, including STEM, humanities, and social sciences, sourced from authentic educational materials and manually validated by native speakers and educators. The dataset includes 10,969 Kazakh questions and 12,031 Russian questions, reflecting Kazakhstan's bilingual education system and rich local context. Our evaluation of several state-of-the-art multilingual models (Llama3.1, Qwen-2.5, GPT-4, and DeepSeek V3) demonstrates substantial room for improvement, as even the best-performing models struggle to achieve competitive performance in Kazakh and Russian. These findings highlight significant performance gaps compared to high-resource languages. We hope that our dataset will enable further research and development of Kazakh-centric LLMs\footnotemark.
\end{abstract}
\footnotetext{\datasetname can be accessed at \url{https://huggingface.co/datasets/MBZUAI/KazMMLU}.}

\section{Introduction} 

\begin{figure}[t]
\centering
\includegraphics[width=0.5\textwidth]{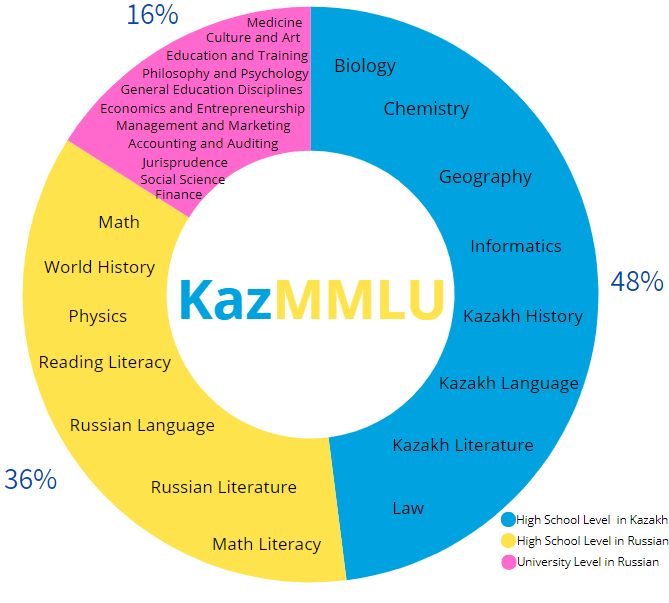}
\caption{Overview of the dataset. This diagram illustrates the distribution of questions by educational level (High School and University) and language (Kazakh and Russian), along with the variety of subjects covered.}
\label{fig:overview}
\end{figure}



With a population exceeding twenty million, the Republic of Kazakhstan in Central Asia remains underrepresented in the field of natural language processing (NLP) \cite{joshi-etal-2020-state}. This gap is highlighted by the limited progress in developing large language models (LLMs) and evaluation benchmarks specifically tailored to the languages and the cultural context of Kazakhstan. 
Kazakh, a Turkic language spoken by more than fourteen million people (around 70\% of the population), holds substantial cultural and geopolitical significance in Central Asia. Russian, used by approximately 15\% of the population, serves as the country’s second primary language.\footnote{\url{https://glottolog.org/}} 

Although Kazakh appears in certain multilingual datasets \citep{yeshpanov2024kazqadkazakhopendomainquestion, yeshpanov-etal-2022-kaznerd, yeshpanov2024kazsandra}, most of these resources rely heavily on translations from English, lacking the cultural richness essential for inclusive LLM development. Previous work has primarily addressed classic NLP tasks such as named entity recognition \cite{yeshpanov-etal-2022-kaznerd} and sentiment analysis \cite{yeshpanov2024kazsandra}. Meanwhile, recent developments in LLM research have shifted toward more reasoning-focused evaluation \cite{meta2024Llama3,openai2024gpt4o}, highlighting a clear research gap for inclusive NLP in the Kazakh context.

Here, we aim to bridge this gap. In particular, we introduce \datasetname, a curated dataset of school- and university-level questions from Kazakhstan, available in both Kazakh and Russian. \datasetname follows the framework of the Massive Multitask Language Understanding (MMLU) dataset \cite{hendrycks2021measuringmassivemultitasklanguage,koto-etal-2024-arabicmmlu, li2024cmmlumeasuringmassivemultitask, koto-etal-2023-large}, which features multiple-choice questions across various subjects and education levels. MMLU has become a standard benchmark for evaluating LLMs' reasoning and knowledge capabilities \cite{meta2024Llama3,gemmateam2024gemma2improvingopen,qwen2025qwen25technicalreport}. Unlike general MMLU, \datasetname incorporates Kazakhstan-specific content, including topics on Kazakh history, traditions, and linguistics, while also reflecting the country's multilingual landscape by providing questions in both Kazakh and Russian.

As shown in Figure \ref{fig:overview}, the dataset is divided into two categories: High School and University. \datasetname consists of approximately 48\% of the questions in Kazakh and 52\% in Russian. The High School section includes questions in both Kazakh and Russian, covering subjects such as Mathematics, Physics, and Kazakh Literature. The University section only features questions in Russian, focusing on professional disciplines such as Law, Economics, and Medicine. This structure aligns with Kazakhstan's bilingual education system and provides a more representative benchmark for evaluating LLMs in the region. \datasetname is sourced from authentic educational materials, including national exams, textbooks, and professional certification repositories. Each question is accompanied by metadata, including the subject, level, source, and correct answer key, ensuring transparency and usability for downstream evaluations.

Our contributions can be summarized as follows:
\begin{itemize}
\item We present the first \textbf{MMLU-style dataset} specifically tailored to the Kazakhstan context, covering diverse subject areas across educational and professional levels. The dataset is made available in both Kazakh and Russian.
\item We evaluate various \textbf{multilingual LLMs}, including Llama3.1 \cite{meta2024Llama3}, Qwen \cite{qwen2025qwen25technicalreport}, GPT-4o \cite{openai2024gpt4o}, BLOOMZ \cite{muennighoff2023crosslingualgeneralizationmultitaskfinetuning}, mT0 \cite{muennighoff2023crosslingualgeneralizationmultitaskfinetuning}, and DeepSeek V3 \cite{deepseek2024}, across different model sizes. 
\item We conduct a thorough analysis of the top-performing open-source models across various dimensions, encompassing (1) individual \textbf{subject areas, educational levels, and Kazakhstan-specific topics}, (2) \textbf{few-shot inference performance}, (3) \textbf{model confidence}, and (4) the \textbf{influence of negation} on model performance. This comprehensive evaluation framework allows us to identify the performance gaps and opportunities for improvement in multilingual LLMs when applied to Kazakh and Russian contexts.
\end{itemize}

\section{Related Work} 

\paragraph{Language Models in Kazakh and Russian}
Prominent models such as OpenAI's ChatGPT, Anthropic's Claude, and Yandex's Yandex-GPT are designed to handle multiple languages, including Russian and Kazakh, enabling a wide range of applications from translation to content generation~\citep{openai2024gpt4o,anthropic_claude, yandex_yandexgpt}. Additionally, open-source models like Meta's Llama series provide multilingual support~\citep{meta2024Llama3}. While these models can produce text in Kazakh, they were not specifically trained or fine-tuned for it. In contrast, the Aya model, an open-access multilingual LLM, supports 101 languages, including Kazakh~\citep{ustun-etal-2024-aya}.

Kazakh-specific language models remain limited, with most multilingual models providing only partial support. To address this, \citet{issai_kazllm} and \citet{koto2025llama} introduced KazakhLLM and Sherkala, both fine-tuned on Kazakh data from Llama. However, their evaluations have largely relied on machine-translated benchmarks.


\paragraph{NLP Benchmark for Kazakhstan Context}


Evaluating LLMs across diverse linguistic and cultural contexts is increasingly critical; however, existing benchmarks overlook Kazakhstan. While benchmarks such as XCOPA~\citep{ponti2021xcopa}, XGLUE~\citep{liang-etal-2020-xglue}, and XTREME~\citep{hu2020extreme} assess cross-lingual performance, they exclude Kazakh, and GlobalMMLU~\citep{cohere2024globalmmlu} lacks Kazakhstan-specific content. Recent efforts like Kardeş-NLU~\citep{senel-etal-2024-kardes} explore cross-lingual transfer for Turkic languages including Kazakh, but focus on NLU tasks specifically, without domain-specific reasoning or local context.
In contrast, several Kazakh-specific datasets exist, including KazNERD~\citep{yeshpanov-etal-2022-kaznerd} for named entity recognition, KazQAD~\citep{yeshpanov2024kazqadkazakhopendomainquestion} for question answering, and KazSANDRA~\citep{yeshpanov2024kazsandra} for sentiment analysis. However, these datasets focus on narrow tasks and do not assess reasoning, factual recall, or domain-specific knowledge. TUMLU~\citep{isbarov2025tumlu} provides a native-language benchmark for Turkic languages including Kazakh, but lacks Kazakhstan-specific context or domain-specific evaluation.
To address these limitations, \datasetname{} presents a large-scale, Kazakhstan-specific benchmark covering STEM, humanities, and social sciences. Unlike previous datasets, \datasetname{} supports a holistic evaluation of reasoning and domain-specific knowledge, offering a more accurate assessment of multilingual LLM capabilities and advancing AI for low-resource languages.

To further illustrate the differences between \datasetname{} and previous benchmarks, we compare it with two existing datasets, SIGTURK~\citep{maxutov-etal-2024-llms} and INCLUDE~\citep{romanou2024include}, in Table~\ref{tab:comparison-mmlu}.
As shown, \datasetname{} is the only dataset that incorporates \textbf{real-world educational materials, professional subjects}, and \textbf{domain-specific reasoning} in both Kazakh and Russian, offering a more \textbf{localized} and \textbf{comprehensive} evaluation of LLMs. This structured assessment underscores \textbf{importance of country-specific benchmarks} in multilingual NLP research and contributes to bridging the gap in Kazakh language understanding.

\begin{table*}[t!]
\centering
\renewcommand{\arraystretch}{1.2} 
\setlength{\tabcolsep}{5pt} 
\footnotesize 
\begin{tabularx}{\textwidth}{X X X X} 
\toprule
\textbf{Feature} & \textbf{KazMMLU} & \textbf{SIGTURK} & \textbf{INCLUDE} \\
\midrule
\bf Public Dataset Size & 23,000 questions in Kazakh and Russian & 3,000 questions exclusively in Kazakh & 23,741 total questions, including 500 in Kazakh \\
\bf Languages Covered & Kazakh, Russian & Kazakh & 44 (including Kazakh) \\
\bf Kazakh-Specific Content & \textbf{Yes}, sourced from local curriculum, national exams & Limited (Kazakh NLP tasks) & Limited \\
\bf Education Levels & High School, University & Not explicitly structured & General education \\
\bf Subjects Covered & STEM, Humanities, Social Sciences, Law, Medicine & QA, MT, causal reasoning & Broad general knowledge \\
\bf Task Type & Bilingual MCQs reflecting real-world knowledge & QA, classification, generative tasks & General MCQs across languages \\
\bf Model Benchmarking & 41 LLMs (GPT-4o, Llama3.1, DeepSeek V3, etc.) & 7 models on Kazakh NLP tasks & Multiple LLMs across 44 languages \\
\bottomrule
\end{tabularx}
\caption{Comparison of \textbf{KazMMLU} with SIGTURK and INCLUDE.}
\label{tab:comparison-mmlu}
\end{table*}

\section{KazMMLU}

In Kazakhstan, the K-12 education system operates in a multilingual setting, with Kazakh as the primary language of instruction in most schools. However, Russian and other languages, such as Uzbek, Uyghur, and Tajik, are also used in specific regions. The curriculum includes core subjects such as mathematics, science, and history, with students required to study both Kazakh and Russian as part of their language education. At the university and professional levels, Russian remains the dominant language of instruction, especially in fields such as law, medicine, economics, and engineering.




\begin{figure}[t]
\centering
\includegraphics[width=\columnwidth]{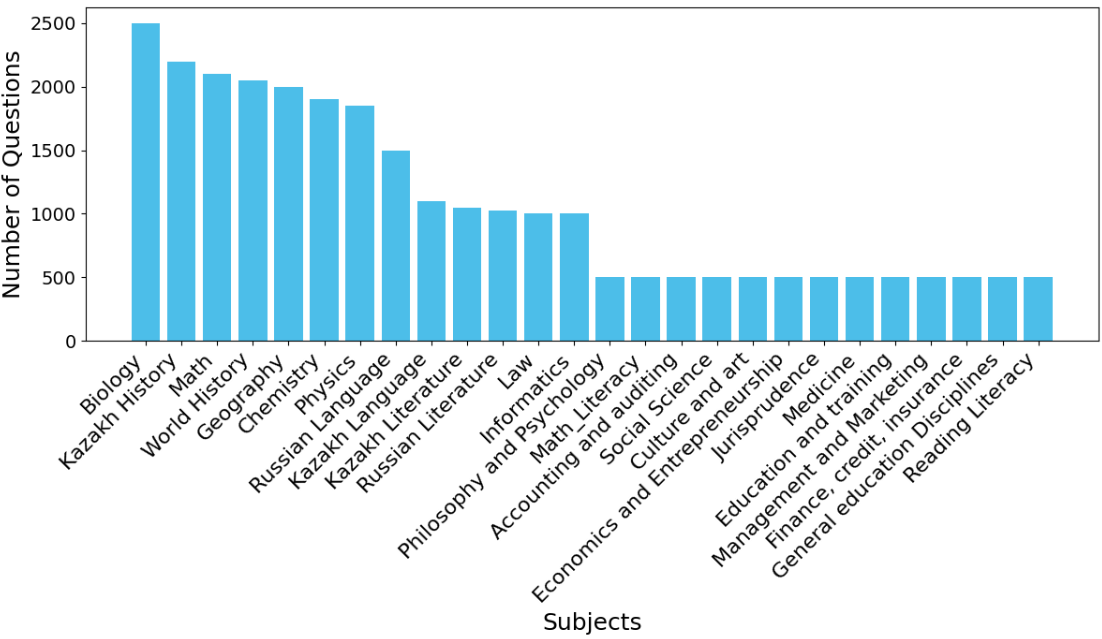}
\caption{Subject-wise distribution of questions in \datasetname{}.}
\label{fig:subject-distributions}
\end{figure}

\datasetname{} spans a broad range of subjects across STEM, humanities, social sciences, and professional studies. As shown in Figure~\ref{fig:subject-distributions}, the dataset has strong coverage in STEM fields, with Biology, Mathematics, and Physics comprising a huge portion of the questions. Humanities and social sciences are also well represented, particularly Kazakh History, World History, and Law, reflecting their central role in the Kazakh educational system.

In addition to academic subjects, \datasetname{} includes questions from professional domains such as Economics, Finance, Jurisprudence, and Medicine, enabling the evaluation of language models in specialized areas. The dataset features content in both Kazakh and Russian, ensuring balanced linguistic representation. All questions are written in Cyrillic script, which remains the prevailing standard in Kazakhstan’s formal education system. This subject and language coverage makes \datasetname{} a robust benchmark for evaluating multilingual language models across diverse domains.

\begin{figure}[t]
\centering
\includegraphics[width=\columnwidth]{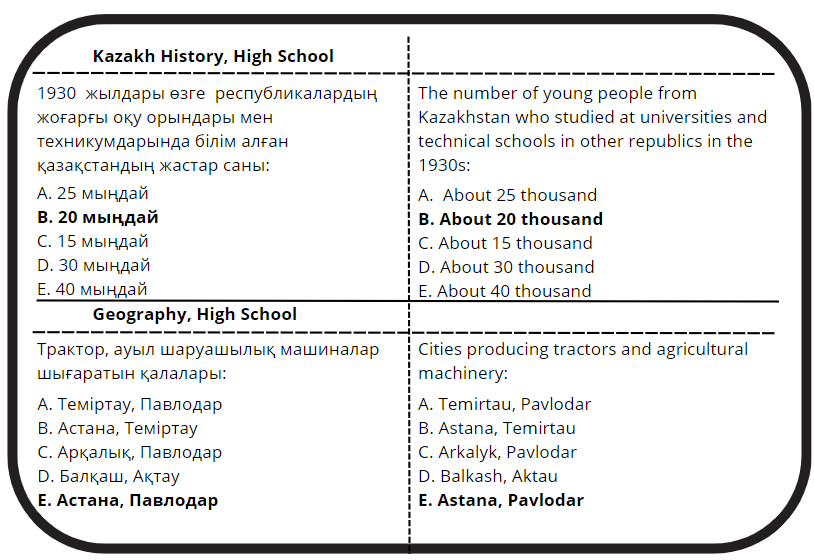}
\caption{Examples of a Kazakh History question and a Geography question from \datasetname{}. The \textbf{left} side shows the original text, and the \textbf{right} side provides the English translation for reference. Correct answers are marked in bold.}
\label{fig:kazakh-mcq}
\end{figure}

To illustrate the question format in \datasetname, Figure \ref{fig:kazakh-mcq} presents a sample multiple-choice question in Kazakh. Answering this question requires an understanding of Kazakhstan's local context, as it covers topics such as history and geography. This example highlights the dataset's diversity across subjects and difficulty levels.

\subsection{Data Construction}

To construct \datasetname, we adopted a systematic approach inspired by MMLU datasets \citep{koto-etal-2023-large,koto-etal-2024-arabicmmlu,li2024cmmlumeasuringmassivemultitask}. The dataset comprises questions sourced from national exams, textbooks, and professional certification materials such as iTest.kz, ymnik.kz, oltest.kz and Book - Shyn Kitap. To ensure diversity, we employed three data collection strategies: (1) automated online crawling, (2) manual transcription from scanned books, and (3) manual extraction from online sources. A detailed breakdown of dataset sources is provided in Appendix~\ref{tab:grouped_dataset_sources}.


For automatic online crawling, we collect question texts, multiple-choice options, correct answer keys, and metadata. For books, authors manually scan materials and apply document processing for machine-readable conversion. Two expert workers fluent in Kazakh and Russian manually extract questions from scanned books and online sources, recording metadata such as source, country, subject, level, and answer key. In total, we compiled 23,000 questions—85\% obtained through automated crawling and the remainder through manual extraction.

Only questions with valid answer keys were included, while multimodal ones requiring images or videos were excluded. For context-dependent questions, annotators ensured necessary context was included. A training workshop clarified guidelines, and weekly check-ins ensured consistency. Annotators were competitively compensated to maintain quality.

\subsection{Quality Control}

Our quality control process primarily targets the automatically-crawled data, as the other two data collection strategies involve direct human involvement. To ensure accuracy, we recruited two professional annotators, each holding at least a bachelor's degree and fluent in both Kazakh and Russian. They manually reviewed all questions to verify correctness and completeness. Any question containing errors or missing components (e.g., incomplete contexts or broken answer options) was discarded. Through this extensive human verification, every question included in \datasetname{} undergoes manual validation, ensuring a high-quality dataset.

Additionally, we developed scripts to detect duplicates, verify metadata, and eliminate errors like duplicate questions, incorrect answer keys, and formatting issues, enhancing dataset reliability



\subsection{Data Statistics}

\datasetname{} comprises 23,000 multiple-choice questions spanning two educational levels: high school and university. As shown in Figure~\ref{fig:overview}, 48\% of the dataset consists of high school-level questions in Kazakh, 36\% in Russian, and 16\% university-level questions in Russian. Table~\ref{tab:subject-areas} outlines the subject distribution, covering STEM, Humanities, Social Sciences, and Languages. Notably, the Humanities, Social Sciences, and Language sections contain extensive Kazakhstan-specific knowledge. The dataset maintains a balanced distribution between Kazakh and Russian, reflecting the bilingual nature of Kazakhstan’s education system.

Table~\ref{tab:average-lengths} presents the average question and answer lengths across educational levels and subject areas. While the overall question length remains relatively consistent between high school and university levels, answer lengths (in characters) tend to be longer at the university level. Additionally, questions in Humanities and STEM subjects are generally longer compared to those in Social Sciences and Languages.


\begin{table}[t]
\centering
    \resizebox{0.8\linewidth}{!}{
    \begin{tabular}{p{1.55cm}p{5.3cm}}
    \toprule
    \textbf{Group} & \textbf{Subjects} \\
    \midrule
    \multirow{5}{*}{Humanities} & Culture and Art (U), Kazakh History (H), Kazakh Literature (H), Philosophy and Psychology (U), Russian Literature (H), World History (H) \\
    \midrule
    \multirow{3}{*}{Language} & Kazakh Language (H), Reading Literacy (H), Russian Language (H) \\
    \midrule
    Other & General Education Disciplines (U) \\
    \midrule
    \multirow{3}{*}{STEM} & Biology (H), Chemistry (H), Informatics (H), Math (H), Math Literacy (H), Medicine (U), Physics (H) \\
    \midrule
    \multirow{9}{*}{\shortstack[l]{Social\\Science}} & Accounting and Auditing (U), Economics and Entrepreneurship (U), Education and Training (U), Finance, Credit, Insurance (U), General Education Disciplines (U), Geography (H), Jurisprudence (U), State and Law (U), Management and Marketing (U), Social Science (U) \\
    \bottomrule
    \end{tabular}
}
\caption{Subject groups covered by KazMMLU. ``H'' indicates high school subjects, and ``U'' indicates university subjects.}
\label{tab:subject-areas}
\end{table}

\begin{table}[t]
\centering
\resizebox{0.8\linewidth}{!}{
    \begin{tabular}{lrr}
    \toprule
    \textbf{Group} & \textbf{Question} & \textbf{Answer} \\
    \midrule
    High School & 78.3 & 16.6 \\
    University & 84.4 & 29.6 \\
    \midrule
    Humanities & 81.3 & 19.1 \\
    Language & 49.3 & 20.3 \\
    Others & 82.0 & 37.1 \\
    STEM & 83.8 & 15.1 \\
    Social Science & 52.9 & 15.4 \\
    \bottomrule
    \end{tabular}
}
\caption{Average question and answer length (in characters) for each educational group and subject area.}
\label{tab:average-lengths}
\end{table}

\begin{figure}[t]
\centering
\includegraphics[width=\columnwidth]{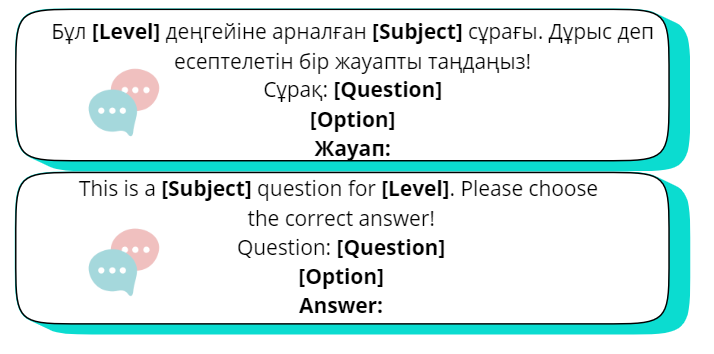}
\caption{Prompt templates in Kazakh and English.}
\label{fig:prompt-example}
\end{figure}

\section{Experiment } 
\subsection{Setup}
We evaluate 27 multilingual LLMs of varying sizes in both zero-shot and few-shot settings. Our selection spans a range of architectures, including BLOOM~\cite{Scao2022BloomA1}, Aya~\cite{ustun-etal-2024-aya}, Mistral~\cite{jiang2023mistral7b}, Gemma~\cite{gemmateam2024gemma2improvingopen}, Qwen~\cite{qwen2025qwen25technicalreport}, GPT-4o~\cite{openai2024gpt4o}, and DeepSeek V3~\cite{deepseek2024}. To ensure regional relevance, we also include models tailored to specific languages: YandexGPT\footnote{\url{yandex.cloud/en/services/yandexgpt}} and Vikhr~\cite{nikolich2024vikhrconstructingstateoftheartbilingual} for Russian, as well as KazakhLLM-8B\footnote{\url{huggingface.co/issai/LLama-3.1-KazLLM-1.0-8B}} and Sherkala-Chat-8B~\cite{koto2025llama} for Kazakh. These regionally fine-tuned models are widely recognized for their focus on local languages.

For the evaluation purpose, we use two distinct prompt configurations to examine the effect of prompt language: (1) a Kazakh prompt with English (Latin script) alphabetic output and (2) an English prompt with English alphabetic output, as illustrated in Figure~\ref{fig:prompt-example}. For placeholders such as \texttt{[Subject]} and \texttt{[Level]}, we use Kazakh in the Kazakh prompt and translate them into English for the English prompt. However, the question and answer choices remain in their original language (Kazakh or Russian). A complete example prompt is provided in Appendix B (Figure~\ref{fig:prompt-examples}).

Following prior studies~\citep{koto-etal-2023-large, li2024cmmlumeasuringmassivemultitask}, we adopt different answer selection methods based on model accessibility. For open-weight models, we apply the \textit{next-token prediction} approach, computing probabilities for each multiple-choice option (A, B, C, D, or E) and selecting the one with the highest probability. This method is well-suited for autoregressive models that perform token-wise scoring. For closed-weight models (e.g., GPT--4o, Yandex-GPT, and DeepSeek V3), we use a \textit{free-text generation} strategy, prompting the model to generate a textual response, from which the predicted answer is extracted via string matching. This approach is necessary due to the lack of direct token probability outputs in closed-source APIs. Regarding inference settings, we used the default configurations provided by their respective APIs or platforms. For evaluation, we use accuracy as the primary metric, following prior studies~\citep{koto-etal-2023-large, li2024cmmlumeasuringmassivemultitask}.





\subsection{Results and Analysis} 

First, we observe that LLMs achieve higher accuracy when prompted in English, as shown in Table~\ref{tab:kazakh-mmlu} and Table~\ref{tab:kazakh-mmlu-kaz}. To provide clearer insights into model performance, we focus on English-prompted results in the main body of the paper.

\begin{table*}[t!]
\setlength{\tabcolsep}{5pt}
\footnotesize
\centering
\begin{tabular*}{\textwidth}{@{\extracolsep{\fill}}lcccccc}
\hline
\textbf{Model} & \textbf{STEM} & \textbf{Social Science} & \textbf{Humanities} & \textbf{Language} & \textbf{Other} & \textbf{Average} \\
\hline
Mistral-7B-Instruct-v0.3            & 37.6 & 47.4 & 43.3 & 30.0 & 40.6 & 41.0 \\
Mistral-7B-v0.3                     & 32.6 & 38.1 & 36.5 & 27.4 & 34.6 & 34.6 \\
\hdashline
Vikhr-Nemo-12B-Instruct-R-21-09-24  & 40.7 & 50.5 & 48.5 & 32.7 & 44.0 & 44.5 \\
\hdashline
aya-23-35B                          & 35.7 & 39.0 & 35.5 & 29.4 & 33.2 & 35.9 \\
aya-23-8B                           & 30.8 & 32.6 & 32.1 & 25.8 & 26.5 & 31.0 \\
\hdashline
Bloom-1.1B                           & 23.6 & 20.7 & 22.1 & 21.6 & 20.8 & 22.1 \\
Bloomz-1.7B                          & 24.0 & 22.4 & 23.5 & 23.0 & 25.8 & 23.4 \\
Bloomz-3B                           & 24.4 & 23.9 & 22.8 & 22.1 & 25.2 & 23.7 \\
Bloomz-7B                           & 23.8 & 24.1 & 23.4 & 23.9 & 22.5 & 23.8 \\
\hdashline
Gemma-2-27B                         & 55.0 & 60.3 & 60.6 & 37.4 & 47.3 & {55.7} \\
Gemma-2-27B-IT                      & 57.3 & 60.5 & 63.2 & 39.1 & 48.3 & {57.4} \\
Gemma-2-9B                          & 51.4 & 57.8 & 55.5 & 36.2 & 42.6 & 52.3 \\
Gemma-2-9B-IT                       & 50.8 & 53.0 & 48.3 & 35.8 & 44.3 & 49.1 \\
\hdashline
KazakhLLM-8B                            & 40.3 & 45.6 & 44.0 & 31.3 & 38.6 & 41.7 \\
\hdashline
Llama3.1-70B                        & 58.0 & 59.1 & 57.4 & 41.8 & 49.3 & {56.2} \\
Llama3.1-70B-instruct               & 51.2 & 48.1 & 51.3 & 33.9 & 45.6 & {48.3} \\
Llama3.1-8B                         & 37.9 & 44.6 & 41.7 & 28.9 & 38.3 & 39.7 \\
Llama3.1-8B-instruct                & 43.0 & 47.8 & 50.6 & 30.4 & 39.3 & 44.6 \\
\hdashline
mt0-large                           & 24.6 & 23.7 & 23.9 & 22.9 & 23.5 & 24.0 \\
mt0-xl                              & 28.2 & 37.0 & 37.7 & 26.5 & 31.9 & 32.8 \\
mt0-xxl                             & 31.3 & 40.5 & 33.0 & 29.5 & 37.9 & 34.4 \\
\hdashline
qwen-2.5-7B                         & 45.2 & 43.1 & 44.1 & 29.4 & 39.9 & 42.5 \\
qwen-2.5-7B-instruct                & 47.5 & 51.0 & 51.4 & 33.9 & 42.3 & 47.8 \\
\hdashline
Sherkala-Chat-8B                  & 43.1 & 49.8 & 50.0 & 34.9 & 38.3 & 45.6 \\
\hdashline
GPT-4o                              & 70.0 & 81.9 & 83.3 & 73.4 & 62.1 & \textbf{76.6} \\
DeepSeek V3                         & 77.6 & 81.3 & 78.9 & 61.2 & 65.1 & \textbf{76.9} \\
YandexGPT                           & 54.8 & 70.6 & 63.7 & 42.6 & 57.0 & 60.2 \\
\hline
\end{tabular*}
\caption{Performance of different models on the Kazakh MMLU benchmark across different subject categories using \textbf{English} prompt. ``Average'' means the average across all subject areas in KazakhMMLU.}
\label{tab:kazakh-mmlu}
\end{table*}

\paragraph{Results across all models}
Table ~\ref{tab:kazakh-mmlu} presents the average accuracy for each subject area across 27 models on the \datasetname{} using English prompts. The performance analysis reveals several notable patterns. GPT-4o and DeepSeek V3 emerge as the top performers, achieving remarkably similar average scores of 76.6\% and 76.9\% respectively, outperforming other models. Gemma-2-27B instruction-tuned model achieved the highest open-source score (57.4\%), followed by Llama3.1-70B (56.2\%) and its base variant (55.7\%). Interestingly, the impact of instruction tuning varies across model families - while Gemma-2-27B-IT showed a slight improvement over its base model (+1.7\%), Llama3.1-70B-instruct performed worse than its base variant (-7.9\%). However, for smaller models, instruction tuning appears more beneficial, as seen in Llama3.1-8B-instruct outperforming its base model by 4.9\% (44.6\% vs 39.7\%).

Consistently across all models, the Language category seems to be the most challenging, with scores lower than other categories. 

\begin{figure}[t]
\centering
\includegraphics[width=0.5\textwidth]{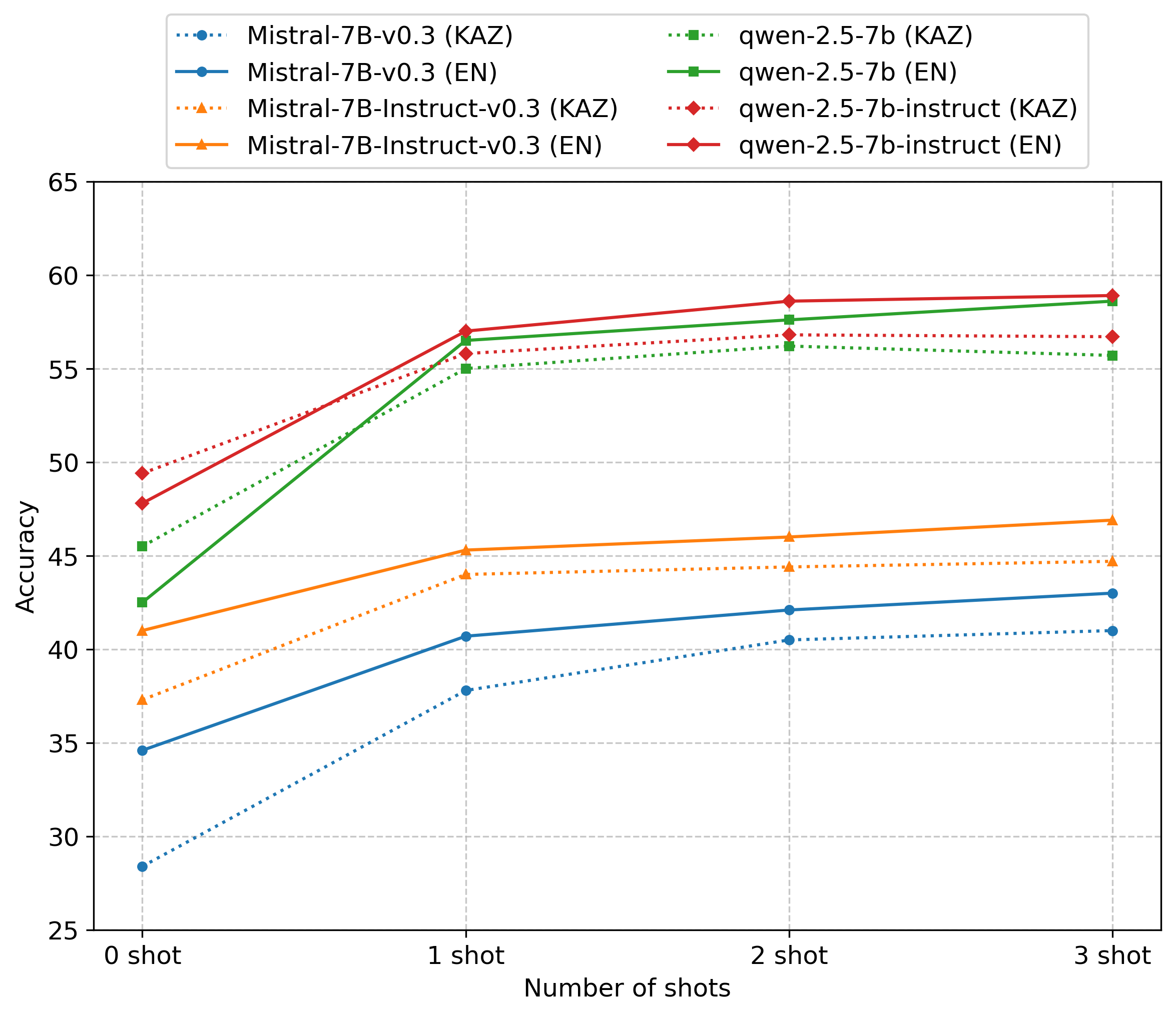}
\caption{The few-shot accuracy (\%) of LLMs on \datasetname{}, averaged across all tasks, comparing base models and instruction-tuned models using Kazakh (dotted lines) and English (solid lines) prompts.}
\label{fig:few-shot}
\end{figure}

\paragraph{Few-Shot Performance} 
As shown in Figure~\ref{fig:few-shot}, our few-shot results show a consistent improvement across all models as the number of shots increases, with Qwen-2.5-7B and Mistral-7B-v0.3 benefiting the most.
English prompts consistently outperform Kazakh prompts in 1, 2, and 3-shot settings, though this trend does not hold in 0-shot, where Kazakh sometimes performs better. Instruction-tuned models also improve, though the gains are smaller, with Qwen-2.5-7B-Instruct (English prompt) increasing from 47.8\% (0-shot) to 58.9\% (3-shot). The largest accuracy jumps occur between 0-shot and 1-shot, indicating that even a single in-context example significantly enhances model understanding. Overall, these results highlight the robustness of few-shot learning across diverse model architectures and prompt settings.

\begin{figure}[t]
\centering
\includegraphics[width=\columnwidth]{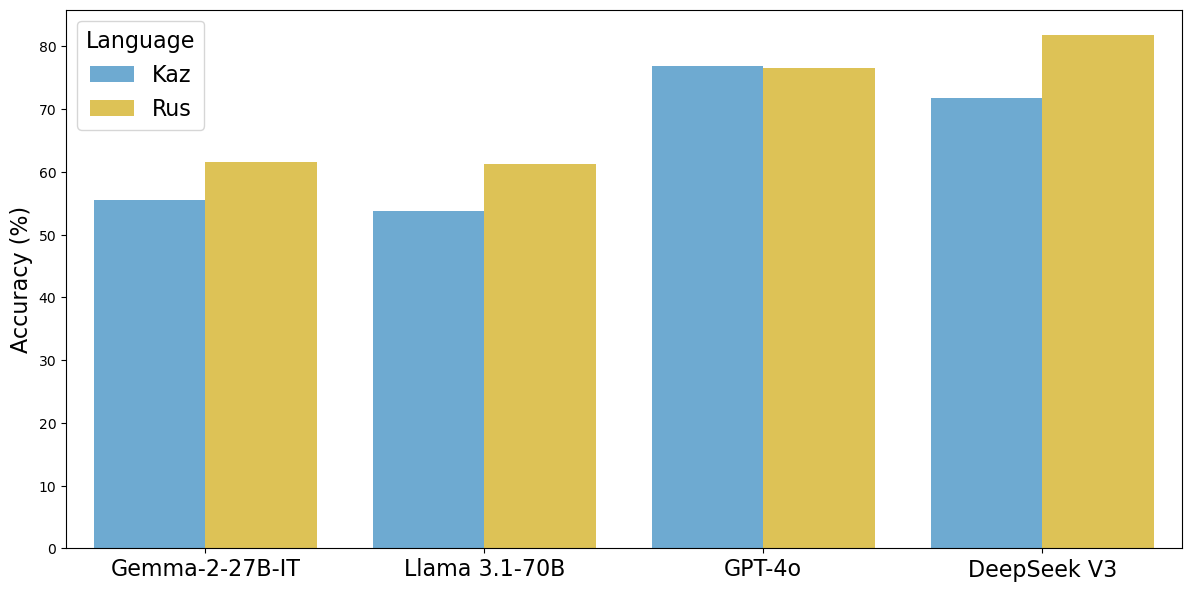}
\caption{LLM Performance across different languages (only at the high school level).}
\label{fig:analysis_language}
\end{figure}

\paragraph{Kazakh vs Russian Performance}
In Figure~\ref{fig:analysis_language}, we compared model performance across two languages: Kazakh and Russian. 
To ensure a fair comparison, we only considered the High School level subjects and excluded the Professional \& University level tasks because they are not available in Kazakh. The results indicate that GPT-4o achieves the highest accuracy in Kazakh, scoring 76.90\%, while DeekSeek performs best in Russian with an accuracy of 81.4\%. 
Llama3.1-70B and Gemma-2-27B-IT show lower but comparable results, with a slight advantage in Russian over Kazakh.
Overall, models tend to perform slightly better in Russian than in Kazakh, which could be due to differences in training data availability, language complexity, or tokenization differences.

\begin{figure}[t]
\centering
\includegraphics[width=\columnwidth]{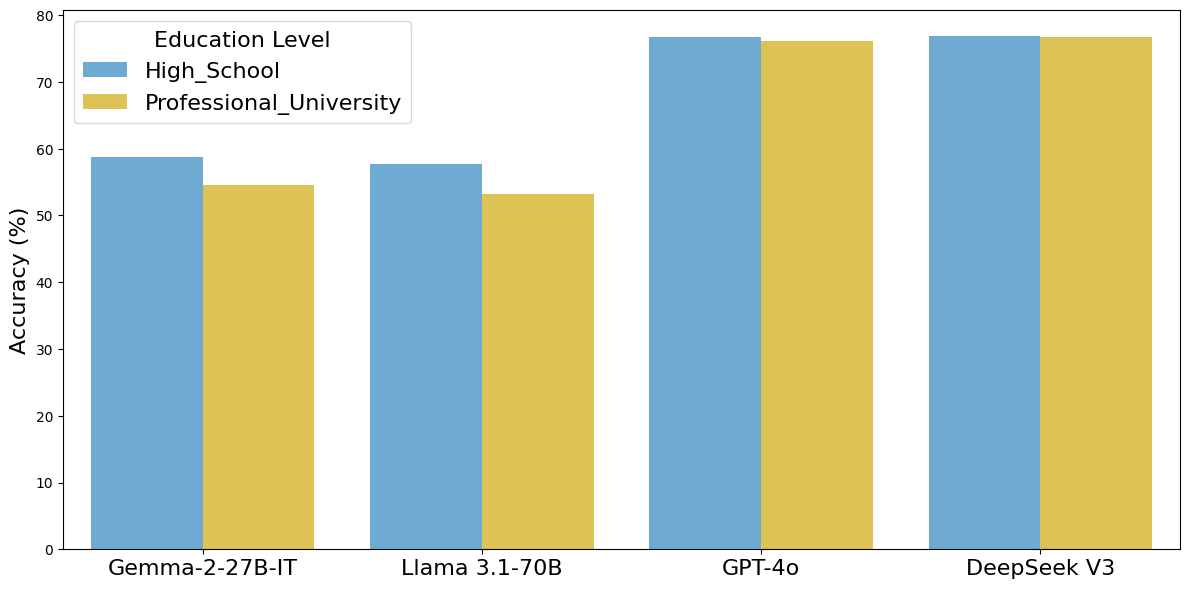}
\caption{LLM Performance across different education levels.}
\label{fig:analysis_level}
\end{figure}

\paragraph{Results Across Education Level}
Results in Figure~\ref{fig:analysis_level} indicate that GPT-4o performs approximately the same across both education levels. 
Similarly, DeepSeek V3 maintains a balanced performance, but with a slight preference towards High School.
In contrast, the open-source models Llama3.1-70B and Gemma-2-27B-IT show considerably lower accuracy and a pronounced gap between education levels. Llama3.1-70B achieves 57.7\% in High School but drops to 53.2\% in Professional \& University. These results suggest that while proprietary models generalize well across different subject complexities, open-source models struggle more with specialized university-level knowledge.

\paragraph{Negation Sensitivity Analysis} Table~\ref{tab:negation_analysis} presents the accuracy of Llama3.1-70B, Gemma-2-27B-IT, and DeepSeek V3 on \textit{negation-sensitive subjects}, comparing performance with and without negation. The results indicate that DeepSeek V3 consistently outperforms both Llama3.1-70B and Gemma-2-27B-IT, demonstrating greater resilience to negation-based reasoning challenges.

\begin{table*}[t!]
\scriptsize
    \centering
    \resizebox{0.8\textwidth}{!}{
    \begin{tabular}{p{4cm}ccc}
        \toprule
        \textbf{Subject} & \textbf{Model} & \textbf{W/o Negation} & \textbf{W/ Negation} \\
        \midrule
        \multirow{3}{*}{\textit{Jurisprudence (University)}} & Llama3.1-70B & \textbf{56.2} & 55.2 \\
        & Gemma-2-27B-IT & 55.2 & \textbf{56.5} \\
        & DeepSeek V3 & \textbf{78.1} & 76.4 \\
        \midrule
        \multirow{3}{*}{\textit{Law (High School)}} & Llama3.1-70B & \textbf{60.8} & 59.0 \\
        & Gemma-2-27B-IT & \textbf{59.0} & {58.1} \\
        & DeepSeek V3 & \textbf{79.5} & 78.1 \\
        \midrule
        \multirow{3}{*}{\textit{Reading Literacy (High School)}} & Llama3.1-70B & 57.1 & \textbf{50.0} \\
        & Gemma-2-27B-IT & \textbf{100.0} & 87.5 \\
        & DeepSeek V3 & \textbf{85.7} & 83.5 \\
        \midrule
        \multirow{3}{*}{\textit{Philosophy and Psychology (University)}} & Llama3.1-70B & 55.9 & \textbf{56.6} \\
        & Gemma-2-27B-IT & \textbf{56.6} & 55.9 \\
        & DeepSeek V3 & \textbf{83.2} & 81.9 \\
        \bottomrule
    \end{tabular}}
    \caption{Model accuracy on negation-sensitive questions across various subjects. Bold values indicate higher accuracy in each category.}
    \label{tab:negation_analysis}
\end{table*}

\begin{figure}[t]
    \centering
    \includegraphics[width=\columnwidth]{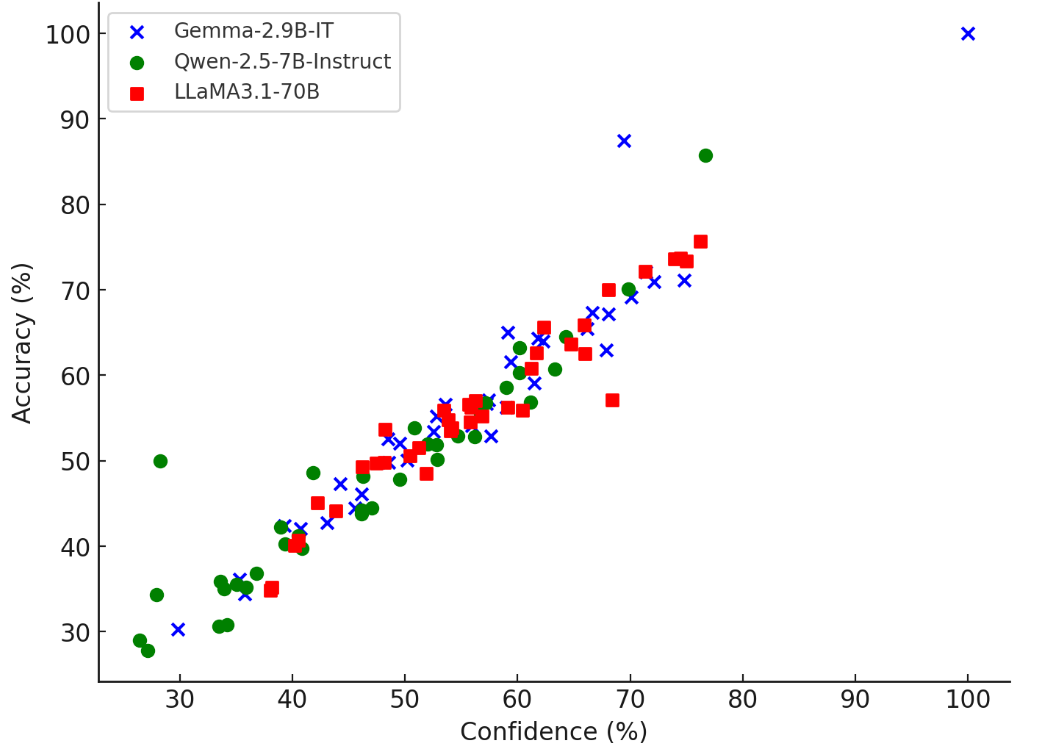}
    \caption{Confidence vs. Accuracy for different models in a zero-shot setting. \textbf{Confidence (\%)} denotes the average probability scores in percentage.}
    \label{fig:confidence-vs-accuracy}
\end{figure}

\begin{figure}[t]
    \centering
    \includegraphics[width=\columnwidth]{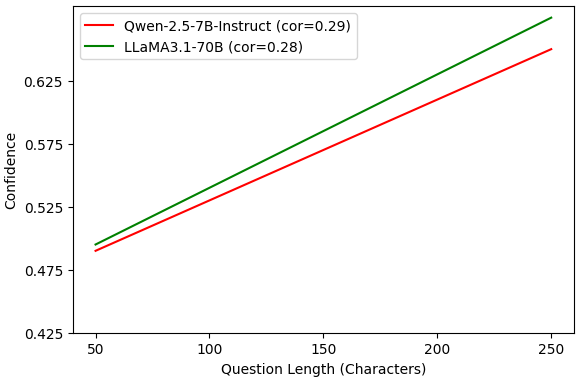}
    \caption{Correlation between model confidence and question length across different models.}
    \label{fig:question-length}
\end{figure}

To systematically analyze the impact of negation, we employed a negation phrase filtering method inspired by ArabicMMLU \cite{koto-etal-2024-arabicmmlu}. Specifically, we identified questions containing common negation phrases in Kazakh: \begin{otherlanguage*}{kazakh}
жоқ (no), емес (is not), болмайды (not allowed), жарамайды (prohibited), невозможно (impossible), не (not), нельзя (forbidden) \end{otherlanguage*}. After applying this filtering, we obtained a total of 2,554 negation-related questions. To validate our filtering accuracy, we randomly sampled 100 questions and manually inspected them. The detection accuracy exceeded 92\%, confirming the reliability of our filtering process.

For Llama3.1-70B and Gemma-2-27B-IT, accuracy generally decreases on questions containing negation, suggesting a negative impact on reasoning capabilities. Notably, Llama3.1-70B exhibits a larger drop in accuracy, particularly in \textit{Reading Literacy}, where performance declines from 57.1\% to 50.0\%. Meanwhile, Llama3.1-70B demonstrates greater robustness in certain cases, showing less fluctuation in performance across negation and non-negation settings.


\paragraph{Model Confidence}
We analyze whether the evaluated models, including Gemma-2.9B-IT, Qwen-2.5-7B-Instruct, and Llama3.1-70B, are well-calibrated when answering \datasetname{} questions by comparing the probability of the correct answers with the actual accuracy for each subject and level combination. The answer probability is obtained through softmax normalization across the available candidate answers. 
In Figure~\ref{fig:confidence-vs-accuracy}, we observe a strong correlation between model confidence and accuracy, with correlation scores exceeding 0.9 across the evaluated models. This indicates that models with higher confidence generally produce more accurate predictions. 


Additionally, we investigate the correlation between model confidence and question length in Figure~\ref{fig:confidence-length}. The results show that question length has minimal impact on model confidence, as evidenced by the weak correlation scores across all evaluated models.
Qwen-2.5-7B-Instruct ($r = 0.29$) and Llama3.1-70B ($r = 0.28$) show a mild positive correlation, suggesting that \textbf{longer questions slightly increase model confidence}. However, the effect is weak overall, indicating that confidence calibration remains relatively stable across different question lengths.

\section{Conclusion and Future Work} 
We introduced \datasetname{}, the first large-scale multi-task language understanding dataset designed to evaluate real-world knowledge in Kazakhstan's bilingual setting. Through experiments with over 23K multiple-choice questions spanning various subjects and education levels, we observed that models perform much better in Russian than in Kazakh, with proprietary models such as GPT-4o and DeepSeek V3 achieving the highest accuracy. 

\datasetname{} provides a \textbf{bilingual} (Kazakh and Russian) evaluation framework tailored to Kazakhstan's educational and professional landscape, distinguishing itself from previous multilingual benchmarks. Unlike SIGTURK, which focused on Kazakh NLP tasks, and INCLUDE, which lacks Kazakhstan-specific content, \datasetname{} enables a localized and comprehensive assessment of LLMs in Kazakhstan. 

Future research directions include extending \datasetname{} to multimodal evaluations, improving reasoning-based question assessments, and mitigating biases in data sources. We hope that this benchmark will encourage further development of high-performance LLMs for Kazakh and other low-resource languages.

\section*{Limitations} 
While we are confident that our benchmark will significantly advance the development of Kazakh LLMs, it is important to acknowledge certain limitations that need to be addressed in future research. We outline these limitations as follows:

\paragraph{Limited Modality} \datasetname{} is focused solely on text-based assessment, and the exploration of multimodal questions (including those involving images, audio, or other media types) has been excluded. Future work could explore the integration of multimodal content to better reflect real-world applications, such as vision-language tasks, speech recognition, and interactive assessments.

\paragraph{Lack of Explicit Reasoning Evaluation} While \datasetname{} provides a broad and representative set of multiple-choice questions, it does not explicitly evaluate reasoning processes beyond answer selection. Investigating how models approach complex reasoning, justification, and open-ended question answering would be a valuable direction for further improvement.

\paragraph{Static Evaluation Limitation} \datasetname{} primarily evaluates static model performance on pre-defined questions, which may not fully capture how models generalize to dynamic, real-world language use. Exploring benchmarks that assess interactive and adaptive reasoning, as well as domain adaptation, could enhance our understanding of model capabilities in evolving contexts.

By addressing these limitations, future research can further refine the evaluation of Kazakh LLMs, ensuring more robust, fair, and practically useful language models for Kazakhstan and beyond.

\section*{Ethics and Broader Impact}

We adhered to the internal policies of web resources while scraping data and included only publicly available information verified by authorities.

All human subjects in our study provided informed consent, were fully aware of the study's objectives, and had the right to withdraw at any time. They were also appropriately compensated as part of their job.

%

\bibliography{acl_latex}

\clearpage
\appendix
\section{Appendix}
\label{sec:appendix}

\subsection{Additional Prompt Examples}
\begin{figure}[h]
    \centering
    \includegraphics[width=1.0\columnwidth]{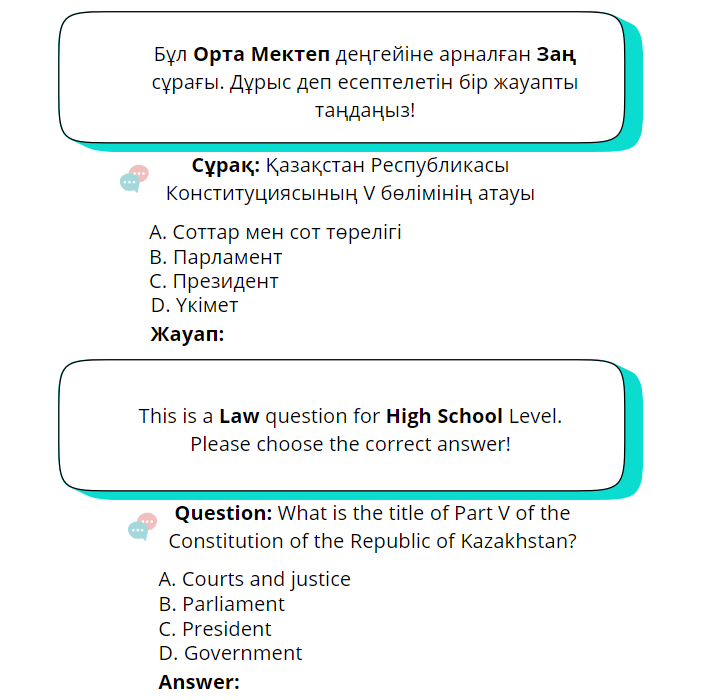}
    \caption{Examples of Kazakh and English prompts. The placeholders are dynamically replaced based on the question context.}
    \label{fig:prompt-examples}
\end{figure}

Figure~\ref{fig:prompt-examples} illustrates the Kazakh and English prompts used in our evaluation. The formatting emphasizes key aspects such as the subject (e.g., \textbf{Law}) and the educational level (e.g., \textbf{High School}). The structure maintains consistency in question presentation, ensuring uniformity across both languages.

\section{Appendix B:}
\label{sec:appendix-b}

\subsection{Additional Multiple-Choice Question Example}
\begin{figure}[h]
\centering
\includegraphics[width=0.48\textwidth]{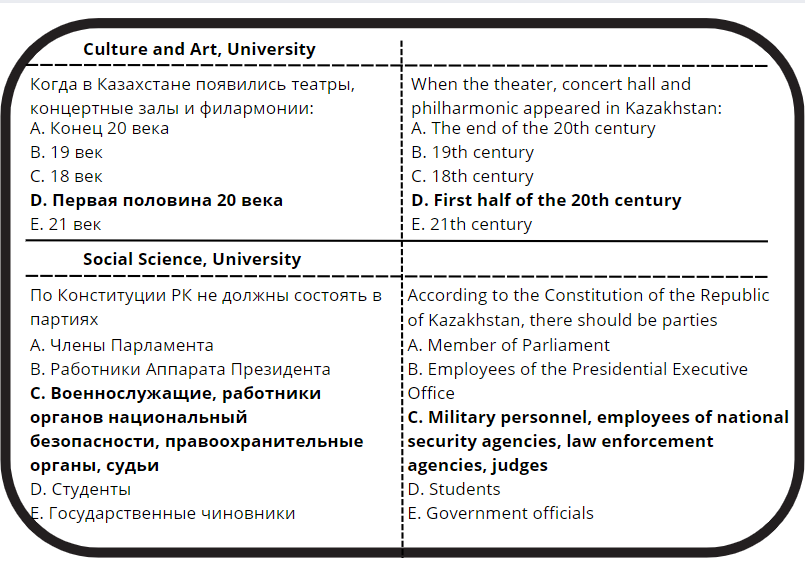}
\caption{Example of a Russian-language multiple-choice question from \datasetname{}. The \textbf{left} side shows the original text, while the \textbf{right} side provides the English translation for illustrative purposes. The bold options represent the correct answer keys.}
\label{fig:russian-mcq}
\end{figure}

In addition to the Kazakh-language question provided in the main document, we present an example of a Russian-language multiple-choice question (Figure~\ref{fig:russian-mcq}). This figure highlights a range of subjects from social sciences to STEM disciplines, demonstrating the dataset’s diversity.

\begin{figure}[t!]
    \centering
    \includegraphics[width=\columnwidth]{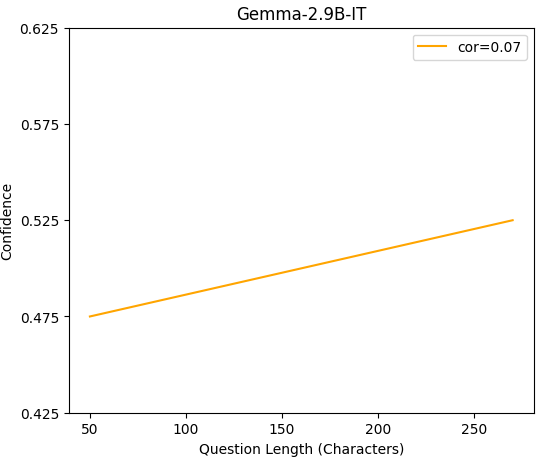}
    \caption{Correlation between model confidence and question length.}
    \label{fig:confidence-length}
\end{figure}

\subsection{Breakdown of Dataset Sources}
Table~\ref{tab:grouped_dataset_sources} provides a categorized breakdown of dataset sources used in \datasetname{}, covering national exams, professional certification tests, and textbooks.

\subsection{Evaluation Results using Kazakh Prompt}
Table~\ref{tab:kazakh-mmlu-kaz} presents the experimental results using Kazakh prompts across various models.

\begin{table*}[h]
\centering
\renewcommand{\arraystretch}{1.9} 
\setlength{\tabcolsep}{5pt} 
\footnotesize
\rowcolors{2}{gray!15}{white} 

\begin{tabularx}{\textwidth}{>{\raggedright\arraybackslash}p{2.5cm}p{2cm}p{2cm}X}
\toprule
\rowcolor{gray!30} 
\textbf{Main Source} & \textbf{Language} & \textbf{Level} & \textbf{Subjects} \\
\midrule
\textbf{itest.kz} & Kazakh & High School &Biology, Chemistry, Geography, Informatics, Kazakh History, Kazakh Literature, Law, Math, Math Literacy, Physics, Reading Literacy, Russian Language, Russian Literature, World History \\
\textbf{oltest.kz} & Russian & University and Professional & Accounting and Auditing, Biology, Culture and Art, Economics and Entrepreneurship, Education and Training, Finance, General Education Disciplines, Jurisprudence, Management and Marketing, Medicine, Philosophy and Psychology, Social Science \\
\textbf{ymnik.kz} & Russian & High School & Biology, Geography, Kazakh History, Kazakh Language, World History \\
\textbf{Book - Shyn Kitap} & Kazakh & High School & Biology, Geography, Kazakh History, Kazakh Language, World History \\
\bottomrule
\end{tabularx}

\caption{Breakdown of dataset sources in \datasetname{}, categorized by domain and subject area. The alternating row colors improve readability.}
\label{tab:grouped_dataset_sources}
\end{table*}

\begin{table*}[!ht]
\setlength{\tabcolsep}{5pt}
\footnotesize
\centering
\begin{tabular*}{\textwidth}{@{\extracolsep{\fill}}lcccccc}
\hline
\textbf{Model} & \textbf{STEM} & \textbf{Social Science} & \textbf{Humanities} & \textbf{Language} & \textbf{Other} & \textbf{Average} \\
\hline
Mistral-7B-Instruct-v0.3  & 34.9 & 43.3 & 38.3 & 27.9 & 31.9 & 37.3 \\
Mistral-7B-v0.3           & 29.5 & 29.3 & 27.7 & 24.6 & 25.2 & 28.4 \\
\hdashline
Vikhr-Nemo-12B-Instruct-R-21-09-24 & 40.3 & 52.7 & 52.7 & 32.8 & 41.9 & 45.9 \\
\hdashline
aya-23-35B                 & 30.5 & 34.1 & 33.0 & 25.8 & 26.5 & 31.5 \\
aya-23-8B                  & 27.2 & 26.9 & 31.2 & 23.2 & 22.8 & 27.4 \\
\hdashline
Bloom-1.1B                  & 23.4 & 19.4 & 20.3 & 22.9 & 19.1 & 21.4 \\
Bloomz-1.7B                 & 22.8 & 20.6 & 23.1 & 23.3 & 20.1 & 22.2 \\
Bloomz-3B                  & 23.2 & 20.6 & 23.6 & 21.9 & 17.8 & 22.2 \\
Bloomz-7B                  & 22.4 & 20.0 & 21.3 & 21.1 & 21.5 & 21.3 \\
\hdashline
Gemma-2-27B                & 57.4 & 63.6 & 64.2 & 41.1 & 46.6 & 58.7 \\
Gemma-2-27B-it             & 59.3 & 62.6 & 63.3 & 40.9 & 46.3 & 58.8 \\
Gemma-2-9B                 & 47.9 & 54.0 & 54.0 & 33.7 & 39.3 & 49.3 \\
Gemma-2-9B-it              & 50.4 & 51.6 & 48.8 & 35.2 & 41.9 & 48.5 \\
\hdashline
KazakhLLM-8B                   & 36.9 & 39.4 & 47.0 & 28.4 & 33.9 & 38.9 \\
\hdashline
Llama3.1-70B               & 58.8 & 64.3 & 64.5 & 44.4 & 50.3 & 59.9 \\
Llama3.1-70B-instruct      & 56.7 & 56.2 & 59.8 & 39.5 & 50.3 & 55.2 \\
Llama3.1-8B                & 34.3 & 39.5 & 38.6 & 28.0 & 31.5 & 36.0 \\
Llama3.1-8B-instruct       & 37.7 & 40.2 & 42.6 & 28.0 & 31.5 & 38.3 \\
\hdashline
mt0-large                  & 24.4 & 24.6 & 22.0 & 22.9 & 24.8 & 23.8 \\
mt0-xl                     & 29.1 & 38.9 & 36.9 & 26.8 & 34.6 & 33.6 \\
mt0-xxl                    & 29.9 & 38.4 & 35.5 & 29.4 & 36.2 & 33.8 \\
\hdashline
Sherkala-Chat-8B         & 36.6 & 39.5 & 43.4 & 30.0 & 34.6 & 38.2 \\
\hdashline
qwen-2.5-7B                & 46.7 & 50.2 & 44.9 & 30.5 & 42.3 & 45.5 \\
qwen-2.5-7B-instruct       & 48.0 & 52.3 & 56.6 & 33.2 & 43.3 & 49.4 \\
\hdashline
GPT4-o                     & 70.5 & 82.1 & 83.2 & 72.7 & 62.1 & 76.7 \\
DeepSeek V3                & 45.4 & 48.0 & 47.8 & 37.4 & 33.3 & 45.5 \\
\hline
\end{tabular*}
\caption{Performance of different models on the Kazakh MMLU benchmark across different subject categories using \textbf{Kazakh} prompt. "Average" means the average across all subject areas in KazakhMMLU.}
\label{tab:kazakh-mmlu-kaz}
\end{table*}

\end{document}